# Facial Expression Recognition Under Partial Occlusion from Virtual Reality Headsets based on Transfer Learning

Bita Houshmand
*Data Science*
*Ryerson University*
*Toronto, Canada*
Email: bita.houshmand@ryerson.ca

Naimul Mefraz Khan
*Electrical and Computer Engineering*
*Ryerson University*
*Toronto, Canada*
Email: n77khan@ryerson.ca

*Abstract*—Facial expressions of emotion are a major channel in our daily communications, and it has been subject of intense research in recent years. To automatically infer facial expressions, convolutional neural network based approaches has become widely adopted due to their proven applicability to Facial Expression Recognition (FER) task. On the other hand Virtual Reality (VR) has gained popularity as an immersive multimedia platform, where FER can provide enriched media experiences. However, recognizing facial expression while wearing a head-mounted VR headset is a challenging task due to the upper half of the face being completely occluded. In this paper we attempt to overcome these issues and focus on facial expression recognition in presence of a severe occlusion where the user is wearing a head-mounted display in a VR setting. We propose a geometric model to simulate occlusion resulting from a Samsung Gear VR headset that can be applied to existing FER datasets. Then, we adopt a transfer learning approach, starting from two pretrained networks, namely VGG and ResNet. We further fine-tune the networks on FER+ and RAF-DB datasets. Experimental results show that our approach achieves comparable results to existing methods while training on three modified benchmark datasets that adhere to realistic occlusion resulting from wearing a commodity VR headset. Code for this paper is available at: https://github.com/bita-github/MRP-FER

*Keywords*-Facial expression recognition; Facial occlusion; Transfer learning; VR

## I. INTRODUCTION

Facial expressions of emotion are a major channel in daily communications to transmit and enhance information not provided by speech [1]. In recent years, machine-based analysis of expressions from human faces has attracted increasing attention as it has broad applications in various areas such as healthcare, emotionally sensitive robots, driver fatigue monitoring, interactive game design and social marketing [2], [3], [4]. However, automatic Facial Expression Recognition (FER) in an unconstrained situation is still difficult. One of the major obstacles for accurate FER outside laboratory environments is partial occlusion in the face [5]. It is very likely that some parts of the face become obstructed by sunglasses, a hat, hands, hair, and the like. In addition, recent advances in technology, specifically in Virtual Reality (VR) has shifted the way we communicate and interact with each other and the environment. However, in a virtual reality setting a large part of the face is occluded by a head-mounted display which blocks facial expressions, there by restricting engagement [6]. Thus, for VR systems to provide rich social interaction, it is vital to be able to recognize and represent these expressions.

Occlusion can substantially change the visual appearance of the face and severely decrease the performance of FER systems. The presence of occlusion increases the difficulty of extracting discriminative features from occluded facial parts due to inaccurate feature location, imprecise face alignment, or face registration error [7]. Types of partial occlusion can be divided into three main categories, temporary, systematic and hybrid [8], [9]. Temporary occlusion occurs when part of the face being obscured temporarily by self-occlusion (e.g., hand or head movement), other objects and change in environmental condition (e.g., illumination and lighting). Systematic occlusion can be caused by the existence of one or more facial components (e.g., hair, mustache, or a scar), or by people using different accessories (e.g. VR headset, glasses, hat, or mask). These types of occlusions are potentially more damaging since they result in whole features of relevance to judging facial expression being obscured. Hybrid occlusion arise in the presence of both systematic and temporary occlusion at the same time.

While lots of studies have been conducted on automatically inferring human expressions from images and video sequences, most of them focused on non-occluded situation and to our knowledge, few of them focused on user's expression in virtual reality environments. Moreover, although the state-of-the-art techniques in FER systems are highly effective for controlled laboratory environments, the existing approaches do not achieve the same accuracy for applications like VR systems in which severe occlusions exist. Furthermore, recently deep learning methods such as Convolutional Neural Networks (CNNs) have outperformed statistical methods. FER performance improvements have been provided by these methods. However training a deep architecture from scratch requires a lot of training data to ensure proper feature learning, and has the difficulty of adjusting many system parameters. Moreover, it needs expensive computational capacity. Existing CNN models are less accurate when handling severe systematic occlusion like VR setting where features of the upper half of the face are completely invisible.

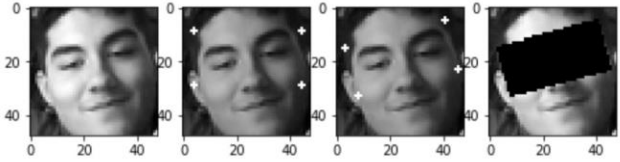

Figure 1. Example of placing a VR patch on a detected face.

In this paper we mainly focus on facial expression recognition in presence of a severe occlusion where the user is wearing a head-mounted display in a VR setting. A unique aspect of the systematic occlusion from arising from a VR headset is that it can be mathematically modeled. Since commodity VR headsets are of known size and shape, we can simulate the occlusion arising from these headsets, rather than trying to collect a new dataset of people wearing such headsets, thus saving significant man hours. In this way, we can employ transfer learning to utilize pre-trained FER networks while simulating occlusion from VR headsets.

The rest of the paper is organized as follows section II reviews related works, section III presents the proposed approach, including our geomtetric model for simulating occlusion and transfer learning-based networks. Section IV describes the experiments. Finally, conclusions are drawn in section V.

## II. RELATED WORKS

Since partial occlusions are a common problem in facial recognition systems and also very frequent in real applications, we review the related approaches considering the similar task to ours which is facial analysis with occluded faces.

Some approaches attempt to minimize occlusion impact by reconstructing occluded parts. Cornejo et al. [10] proposed a methodology which is robust to occlusions through the Weber Local Descriptor (WLD). They used RPCA technique for reconstructing occluded facial expressions over the training set and projected all testing images into the space created by RPCA. The WLD descriptor is applied over the entire facial expression image for extracting textural features. For each feature vector they applied feature dimensionality reduction techniques, such as PCA and LDA sequentially and used SVM and K-NN for classification. Wang et al. [11] proposed a novel framework for FER under occlusion by fusing the global and local features. In global aspect, information entropy is employed to locate the occluded region and principal Component Analysis (PCA) method is adopted to reconstruct the occlusion region of image. Then, the occluded region replaced with the corresponding region of the best matched image in training set and Pyramid Weber Local Descriptor (PWLD) feature is extracted. At last, the outputs of SVM are fitted to the probabilities of the target class by using sigmoid function. For the local aspect, an overlapping block-based method is adopted to extract WLD features, and each block is weighted adaptively by information entropy, Chi-square distance and similar block summation methods are then applied to obtain the probabilities which emotion belongs to. Finally, fusion at the decision level is employed for the data fusion of the global and local features based on Dempster-Shafer theory of evidence.

Deep Convolutional Neural Networks have been pushing the frontier of face recognition over past years. To address the occlusion issue, Li et al. [12] proposed end-to-end trainable Patch-Gated Convolution Neutral Network (PG-CNN), a CNN with attention mechanism which tries to focus on different regions of the facial image and weighs each region according to its unobstructed-ness. It decomposes the feature maps of the whole face to multiple sub-feature maps to obtain diverse local patches and encoded them as a weighed vector by a patch-gated unit using attention net considering its unobstructed-ness. Both local and global representations of facial features are concatenated to serve as a representation of the occluded face, which would alleviate the impact caused by the lack of local expression information. Another approach to improve the ability of classification in occluded condition is transferring features. Xu et al. [13] developed a hybrid model based on deep convolutional networks to increase the robustness of transfer features from deep models. They built a deep CNN composed of four convolutional layers with max-pooling to extract features, followed by a multi-class SVM for emotion classification. To increase robustness to occlusion further, they improve model by merging high-level features of two trained deep ConvNets with the same structure, one is trained on non-occluded images, and the other is trained on same database with additive occluded samples. Non-occluded facial expression classifier can be used as a guidance to facilitate the process of an occluded facial expression classifier. Pan et al. [14] built two deep convolutional neural networks with the same architecture, these two networks are first pretrained with the supervised multi-class cross entropy losses. After pretraining, parameters of non-occluded network are fixed, and the occluded network is further fine-tuned under the guidance of non-occluded network. To incorporate guidance into the network during training, they introduce similarity constraint and loss. Also a mask learning strategy can be adapted to find and discard corrupted feature elements. Song et al. [15] designed a Pairwise Differential Siamese Network (PDSN) to establish a mask dictionary by exploiting the differences between the top convolutional features of occluded and non-occluded face pairs. Each item of this dictionary captures the correspondence between occluded facial areas and corrupted feature elements, which is named Feature Discarding Mask (FDM). When dealing with a face image with random partial occlusions, they generate its FDM by combining relevant dictionary items and multiply it with the original features to eliminate those corrupted feature elements from recognition. The framework proposed in [22] utilizes an efficient filtering method to reduce the search space of face retrieval to increase scalability while keeping representative occluded faces within the search space.

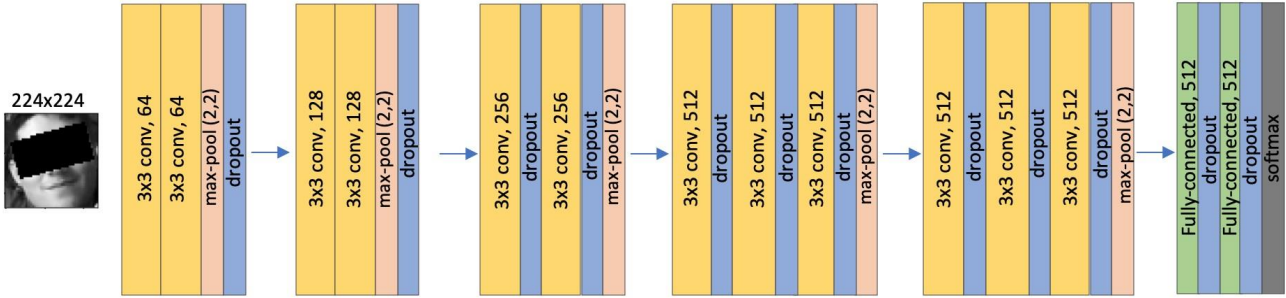

Figure 2. Custom VGG-Face network architecture

To our knowledge just two studies focused on FER in VR setting. Georgescu and Ionescu [16] proposed a model for recognizing the facial expression of a person wearing a virtual reality (VR) headset which essentially occludes the upper part of the face. They trained VGG-f and VGG-face models, on modified training images in which the upper half of the face is completely occluded. This forces the neural networks to find discriminative clues in the lower half of the face. They proceed by fine-tuning the networks in two stages. In the first stage they fine-tune the CNN models on images with full faces. In the second they further fine-tune the models on images in which the upper half of the face is occluded. Hickson et al. [17] also considered facial expression recognition in VR setting. They proposed an algorithm to infer expressions by analyzing only the images of a person's eyes captured from an eye-tracking enabled VR head-mounted display through an IR gaze-tracking camera. In terms of taking user's face images, as not all VR headsets include eye-tracking sensors, we adopt similar approach to [16], considering external camera in VR system which is more generalizable and cost-effective. However, a key limitation in the approach in [16] is that they simply covered the upper half of the face without taking into consideration realistic occlusion arising from VR headsets, which we attempt to achieve in this work.

## III. METHODOLOGY

### A. Simulation of partial occlusion

In a virtual reality setting, the upper half of the face is occluded by a VR headset. Since, there are no public standard occlusion face images database containing systematic occlusion or persons wearing VR headset, we established VR-occluded images by masking upper region on the standard facial expression images. To simulate the occlusion, face detection is applied on gray-scale images based on a modification to the standard Histogram of Oriented Gradients and Linear SVM-based method for object detection. In order to estimate 68 coordinates of facial landmarks that map to facial structure, we followed the approach described in [18] trained on the iBUG 300-W face landmark dataset [19]. For applying VR patch, we initialize VR dimensions to $207.1 \times 98.6 mm$ based on the Samsung Gear VR headset. Typically, a resized set of training images have faces filling various percentages of the image, and it is unlikely that a fixed-sized covering mask fits all the training images. To uniformly scale the VR patch on the training images, we use the distance between the two temporal bones of the facial landmarks as a reference length. Then, we generate the polygonal occluding patch by setting the midpoint of the line passing through eye centre points as the centre coordinate of the VR headset. Moreover, to account for face rotations, we align the resized patch with the axis running through eye centres to reflect accurate simulation of a VR headset on various situations. We obtain the angle of incline by determining the inverse tangent function of changes in y-coordinates to changes in x-coordinates of eye centre points, and then we utilize the rotation matrix to rotate corner points of the blocking patch about its central pivot point on the coordinate plane accordingly. Figure 1 demonstrates the occlusion simulation process. This geometric model provides a more realistic occlusion resulting from wearing a VR headset, rather than simply covering the upper half of the face, as done in [16].

### B. Transfer Learning

In CNNs, images from different datasets share similar low-level features after the convolution process. As it is costly to train a network from scratch, especially on a small-scale dataset, an alternative training strategy for a new dataset is utilizing parameters transferred from pre-trained models and fine-tuning them on the basis of the new dataset. Thus, for classifying facial expressions under severe occlusion we choose two popular architectures, VGG and ResNet pre-trained on a face recognition task which is a closely related to ours.

Table I
MODELS AND CORRESPONDING ACCURACIES

| Model | Dataset | Accuracy (%) |
|---|---|---|
| VGG-Face (from scratch) | FER+ | 65.68 |
| ResNet50 (from scratch) | FER+ | 54.43 |
| VGG-Face (transfer learning) | FER+ | 79.98 |
| ResNet50 (transfer learning) | FER+ | 79.90 |
| VGG-Face [16] | FER+ | 82.28 |
| VGG-Face (transfer learning) | AffectNet | 50.13 |
| ResNet50 (transfer learning) | AffectNet | 47.35 |
| VGG-Face [16] | AffectNet | 49.23 |
| VGG-Face (transfer learning) | RAF-DB | 73.37 |
| ResNet50 (transfer learning) | RAF-DB | 74.76 |

*1) VGG-face:* VGG-face [20] is a convolutional neural network, trained on the VGG-face dataset. It comprises of 13 convolutional layers, each containing a linear operator followed by one or more non-linearities such as ReLU and max pooling. Last three blocks are fully connected followed by a softmax layer of 2622 units for classification.

*2) ResNet:* The residual neural network [21] implements the identity shortcut connection concept or residual block that skips one or more layers. ResNet won the first place on the ILSVRC 2015 classification task. ResNet follows VGG's convolutional layer design. The residual block has two 3 × 3 convolutional layers with the same number of output channels. Each convolutional layer is followed by a batch normalization layer and a ReLU activation function. Then, it skips these two convolution operations and add the input directly before the final ReLU activation function. Generally, the residual mapping is often easier to optimize in practice.

we proceed our work by fine-tuning the models on occluded images generated through applying our geometric simulation model as described before. The custom VGG-face has 13 convolutional and 4 max-pooling layers as VGG-face, interleaved with dropout layers. More specifically, starting from second block, after each convolutional layer, a dropout layer is added with rate starting from 0.1 to 0.55. The dropout layers are effective in avoiding the model to over fit. After all the convolution layers, 2 dense layers are added, each with 512 hidden nodes, followed by a 50% dropout layer. The final dense layer is followed with a soft-max layer consisting of 8 nodes to generate the output. In the ResNet-50 architecture, layers pre-trained on VGGFace dataset are transferred to our deep CNN model, followed by 2 dense layers with 512 hidden nodes and replacing the last 1000 fully connected softmax layer by a 8 fully connected softmax layer. Figure 2 demonstrates the customized architechture of VGG-face.

## IV. EXPERIMENTAL RESULTS

### A. Datasets

To evaluate our models, we use three popular facial expression datasets, FER+, RAF-DB, and AffectNet. These datasets cover different scales of face images and contain images with temporary occlusions.

*1) FER+:* The FER+ dataset [23] is derived from FER2013 which is a large-scale and real-world dataset. It contains 8 emotional classes including anger, disgust, fear, happiness, neutral, sadness, surprise and contempt. Comparing to FER2013, some of the original images are relabeled, while other images, e.g. not containing faces, are completely removed. All face images in the dataset are aligned and resized to 48 × 48. The dataset consists of 28,709 training images, 3,589 validation images and 3,589 test images.

*2) RAF-DB:* Real-world Affective Faces Database [24] is an in-the-wild dataset which contains 30000 facial images posted in social networks and annotated by 40 annotators. The dataset includes 7 classes of basic emotions including surprised, fearful, disgusted, happy, sad, angry and neutral and 12 classes of compound emotions. In our experiment, only images with basic emotions were used, including 12,271 images as training data and 3,068 images as test data.

*3) AffectNet:* The AffectNet dataset [25] is one of the largest annotated datasets which contains more than one million images. It has 450000 manually annotated images with 8 basic emotional classes as FER+ and 3 other categories related to the intensity of valence and arousal including none, uncertain and non-face.

### B. Training

During training, the input to our CNNs is set to a fixed sized 224 × 224 image. We rescaled all training, validation and test images to 224 × 224 pixels. All images are normalized by min-max normalization and the models are trained using data augmentation, which is based on including horizontally flipped images. The training is carried out by optimising the multinomial logistic regression objective using mini-batch gradient descent based on back-propagation with momentum. The batch size is set to 64 and momentum is set to 0.9. The training was regularised by weight decay, the L2 penalty multiplier set to $5 \times 10^{-4}$ and the ratio for dropout regularisation was set to 0.5. The learning rate was initially set to $10^{-2}$, and then decreased by a factor of 10 when the validation set accuracy stopped improving. In case of the ResNet50 architecture we also regularised training by setting max-norm kernel constraint.

### C. Results

For each model, we train our network on each dataset separately with 5-fold cross-validation and report the accuracy numbers in Table I. It can be seen from the table that training models from scratch provides less accurate results in comparison with transfer-learning based approaches. The VGG-Face architecture pre-trained on VGG-Face dataset and fine-tuned on FER+ achieved accuracy of 79.98% which is the highest among the tested models in this paper. comparing the results with the results reported in [16], we see that we provide slightly better accuracy on AffectNet and slightly worse on FER+. However, it is important to note that our geometric occlusion simulation model adheres to more realistic occlusion resulting from a commodity VR headset (Samsung Gear VR), while in [16] the occlusion was generated by simply covering the upper half of the face. So these results are not directly comparable, however it shows that a simulated occlusion with transfer learning can provide promising results by utilizing existing benchmark datasets.

## V. Conclusion

In this paper, we propose a transfer learning-based method for addressing the facial expression recognition in presence of severe occlusion where the user is wearing a head-mounted display in a VR setting. We modified three existing benchmark datasets to mimic VR occlusion by utilizing a geometric simulation model based off of the Samsung Gear VR headset. We test two popular architectures, VGG and ResNet pretrained on the VGG-Face database and further fine-tune parameters of the models on the occluded images. We evaluate our method on FER+, AffectNet and RAF-DB datasets. Our method achieves performance comparable to the state-of-the art results. For future work, we will study how to integrate approaches using external cameras and methods utilizing internal cameras in VR system to improve the accuracy further and provide a rich engagement experience for virtual reality users.